\documentclass{ecai} 

\usepackage{tabularx}
\usepackage{array}
\newcolumntype{M}[1]{>{\centering\arraybackslash}m{#1}} 
\usepackage{latexsym}
\usepackage{amssymb}
\usepackage{amsmath}
\usepackage{amsthm}
\usepackage{booktabs}
\usepackage{enumitem}
\usepackage{graphicx}
\usepackage{color}
\usepackage{hyperref}
\hypersetup{
    pdfborder={0 0 0} %
}
\usepackage{multirow}
\usepackage{booktabs}   %
\usepackage[footnotesize]{caption}
\newtheorem{theorem}{Theorem}

\newcommand{\BibTeX}{B\kern-.05em{\sc i\kern-.025em b}\kern-.08em\TeX}

\begin{document}


\begin{frontmatter}


\paperid{8341} 


\title{GEA: Generation-Enhanced Alignment for\\
Text-to-Image Person Retrieval}

\author[1]{\fnms{Hao}~\snm{Zou}}
\author[2]{\fnms{Runqing}~\snm{Zhang}}
\author[2,1]{\fnms{Xue}~\snm{Zhou}\thanks{Corresponding author: zhouxue@uestc.edu.cn}}
\author[2]{\fnms{Jianxiao}~\snm{Zou}}

\address[1]{School of Automation Engineering, University of Electronic Science and Technology of China}
\address[2]{Shenzhen Institute for Advanced Study, University of Electronic Science and Technology of China}


\begin{abstract}
Text-to-Image Person Retrieval (TIPR) aims to retrieve person images based on natural language descriptions. Although many TIPR methods have achieved promising results, sometimes textual queries cannot accurately and comprehensively reflect the content of the image, leading to poor cross-modal alignment and overfitting to limited datasets. Moreover, the inherent modality gap between text and image further amplifies these issues, making accurate cross‑modal retrieval even more challenging.
To address these limitations, we propose the Generation-Enhanced Alignment (GEA) from a generative perspective. GEA contains two parallel modules: 1) Text-Guided Token Enhancement (TGTE), which introduces diffusion-generated images as intermediate semantic representations to bridge the gap between text and visual patterns. These generated images enrich the semantic representation of text and facilitate cross-modal
 alignment. 2) Generative Intermediate Fusion (GIF) module, which combines cross-attention between generated images, original images, and text features to generate a unified representation optimized by triplet alignment loss.
We conduct extensive experiments on three public TIPR datasets, CUHK-PEDES, RSTPReid, and ICFG-PEDES, to evaluate the performance of GEA. The remarkable results justify the efficacy of our method. More implementation details and extended results are available at {https://github.com/sugelamyd123/Sup-for-GEA}.

\end{abstract}

\end{frontmatter}


\section{Introduction}
Text-to-Image Person Retrieval (TIPR)~\cite{Jiang_2023_CVPR, Qin_2024_CVPR, shu2022see, Qin_2025_CVPR} aims to retrieve specific person images from a large-scale dataset based on a given textual description. By bridging visual and linguistic modalities, TIPR has great potential in practical applications such as public safety and missing person search~\cite{pei2022multi,yaghoubi2021sss}, making it receive widespread attention from both academia and industry. However, despite recent progress, TIPR is still faced with challenges such as the inherent gap between visual and textual modalities. One of the core reasons lies in the discrepancy between the nature of information provided by images and texts: images are rich in low-level visual details such as color, clothing texture, and accessories, while textual descriptions are often abstract, sparse, and semantically ambiguous. This semantic asymmetry makes it difficult to establish accurate correspondence between the above modalities.

Existing methods~\cite{ding2021semantically,li2017person,zhang2018deep} always adopt complex architectures to achieve both global and local semantic alignment between image and text 
\begin{figure}[t]
\centering
\includegraphics[width=8cm]{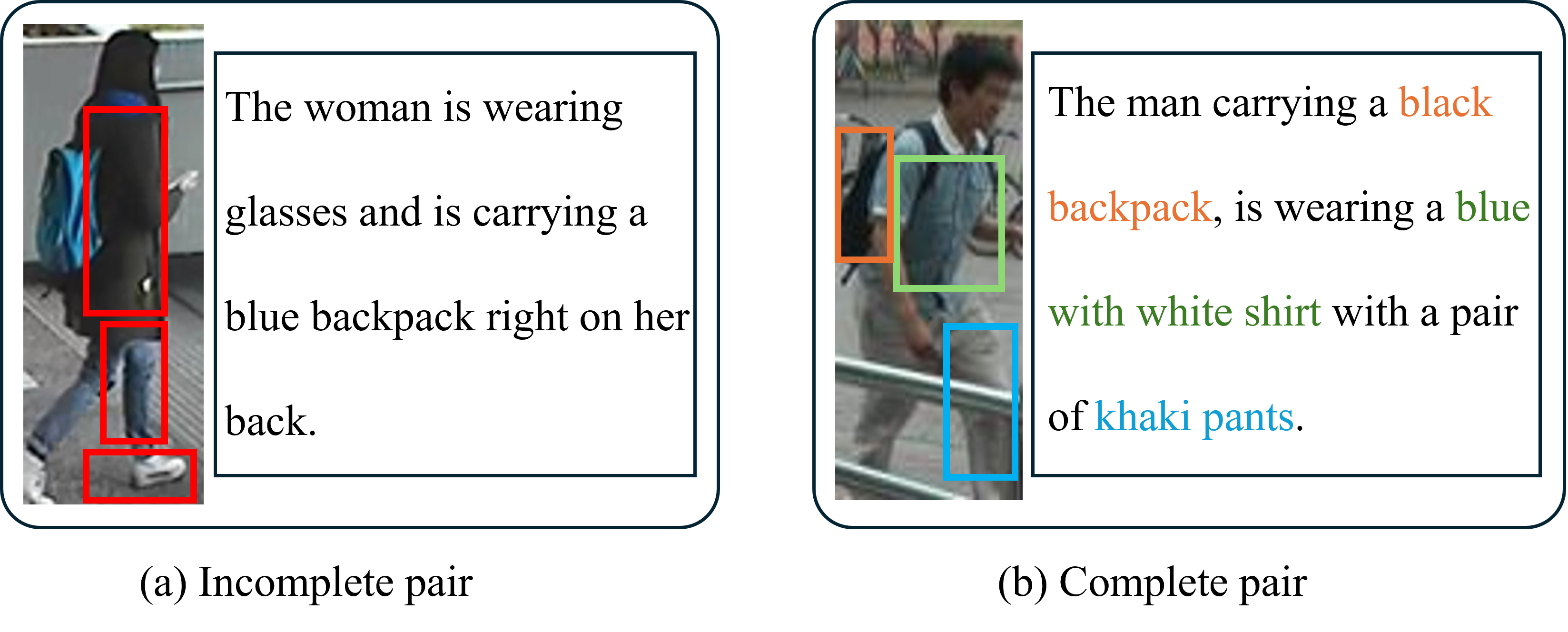}
\caption{Comparison between incomplete and complete textual descriptions.
(a) shows a person image with multiple details highlighted in red boxes (such as coat, jeans, shoes), which were not reflected by incomplete text query, while (b) presents a more complete description. This clearly illustrates how the lack of complete textual exacerbates the modality gap between visual and textual modalities in real-world TIPR scenarios.}
\label{fig:eurai}
\vspace{0.5cm} 
\end{figure}
modalities. With the emergence of CLIP~\cite{radford2021learning}, some TIPR methods~\cite{chen2023towards, Ni_2023_ICCV} have explored leveraging its powerful cross-modal alignment capabilities. These approaches typically adopt CLIP as the backbone and train it on retrieval-specific datasets, achieving impressive results due to inheriting rich visual-linguistic priors from large-scale pretraining. Although these approaches have achieved promising performance, this task still faces two key challenges:
(1) Textual queries often lack the fine-grained details required to accurately describe target images. As illustrated in Figure~1, incomplete pair makes it nearly impossible to establish a one-to-one correspondence between textual descriptions and the visual content of target images, especially subtle visual cues are critical for matching.
(2) Due to privacy concerns and data collection constraints, person image datasets are often limited in size and diversity. Under such conditions, existing methods are prone to overfitting to coarse-grained or simplistic text, failing to capture comprehensive fine-grained information.

To overcome these limitations, we propose a novel approach called Generation-Enhanced Alignment (GEA), which tackles the challenges of incomplete textual descriptions and limited data diversity in TIPR from a generative perspective. Leveraging the powerful generative capabilities of the Stable Diffusion (SD) model~\cite{esser2024scaling}, our method utilizes text-guided image generation to faithfully enrich and extend the semantic content conveyed by textual queries. Even when the input text provides only sparse or limited descriptions, the generated images can capture and amplify the key semantic cues~\cite{rombach2022high}, serving as effective intermediate representations. GEA introduces these diffusion-generated images as semantic bridges between abstract textual descriptions and dense visual content of person images, thereby narrowing the modality gap and facilitating more accurate feature alignment. Furthermore, by enriching the semantic space through generative representations, GEA indirectly enhances model generalization under low-data regimes without requiring additional real-world samples.

Compared with existing methods~\cite{shao2023unified, Zhang_2018_ECCV, qin2022deep} that rely on complex architectures to directly align image-text semantics, our approach leverages the generative power of diffusion models to create intermediate images that serve as semantic bridges. Rather than requiring complete textual descriptions, GEA utilizes diffusion-generated images to amplify and enrich the semantic information associated with a few critical words. These generated images bridge the gap between sparse textual inputs and dense visual content, significantly enhancing cross-modal alignment even under incomplete supervision.
The main contributions of this paper are summarized as follows:

\begin{itemize}
\item[$\bullet$] We introduce the generate method into Text-to-Image Person Retrieval (TIPR). To the best of our knowledge, this is one of the first works to use text-generated images as a "mediator expression" to bridge the modal gap between images and text in TIPR task. 
\item[$\bullet$] We propose a novel fusion structure GEA. To bridge modality differences, we design the Text-Guided Token Enhancement (TGTE) module to enhance text. Through the Generative Intermediate Fusion (GIF) module, our method achieves alignment enhancement of image text at the semantic level.

\item[$\bullet$] Experimental results show that GEA significantly improves TIPR accuracy and generalizes well across multiple datasets, demonstrating the effectiveness and superiority of our approach.
\end{itemize}


\section{Related Work}
We review prior work in two categories: (1) Text-to-Image Person Retrieval (TIPR), a task that explores various strategies to align textual and visual modalities; (2) Text-guided diffusion models, which focus on generating high-quality images conditioned on textual inputs.
\subsection{Text-to-Image Person Retrieval}
Text-to-Image Person Retrieval (TIPR) aims to retrieve images of specific individuals based on textual descriptions. Typically, existing methods~\cite{Zhang_2018_ECCV, Jing_Si_Wang_Wang_Wang_Tan_2020} map both image and text data into a shared embedding space, and the similarity between their features is used to perform retrieval. Early approaches~\cite{chen2018improving, faghri2017improving} mainly focused on encoding both modalities into fixed-length vectors to achieve global feature alignment. Although these methods achieve decent performance in coarse-grained matching scenarios (e.g., identifying body shape or dominant clothing color), they often fail to capture subtle visual details critical for precise retrieval, particularly when dealing with fine-grained attributes such as clothing style or accessory positioning.
Recent works~\cite{shao2023unified, zhao2021weakly, zheng2020dual} have explored fine-grained alignment strategies by associating local image regions with specific words or phrases in text. With the emergence of powerful vision-language models like CLIP~\cite{radford2021learning}, some approaches~\cite{Jiang_2023_CVPR, Qin_2024_CVPR} have attempted to leverage pretrained cross-modal alignment capabilities and further fine-tune them for TIPR. 

The aforementioned methods have made significant progress, they are primarily based on the assumption that the given text description is complete and faithfully reflects the visual content.
However, in real-world scenarios, textual queries are sometimes incomplete, providing only sparse descriptions that omit some key visual details. This semantic insufficiency makes it difficult for models to establish precise one-to-one mappings between text and images, especially for fine-grained attributes. Moreover, due to privacy constraints and annotation costs, person image datasets remain limited in size and diversity. These challenges hinder the ability of current models to robustly learn cross-modal semantics and generalize well, particularly in data-scarce environments.
These limitations motivate the need for a generative-based solution that can enrich textual semantics, bridge the modality gap, and enhance retrieval robustness even under incomplete supervision.

\begin{figure*}[t]
\centering
\includegraphics[width=16cm]{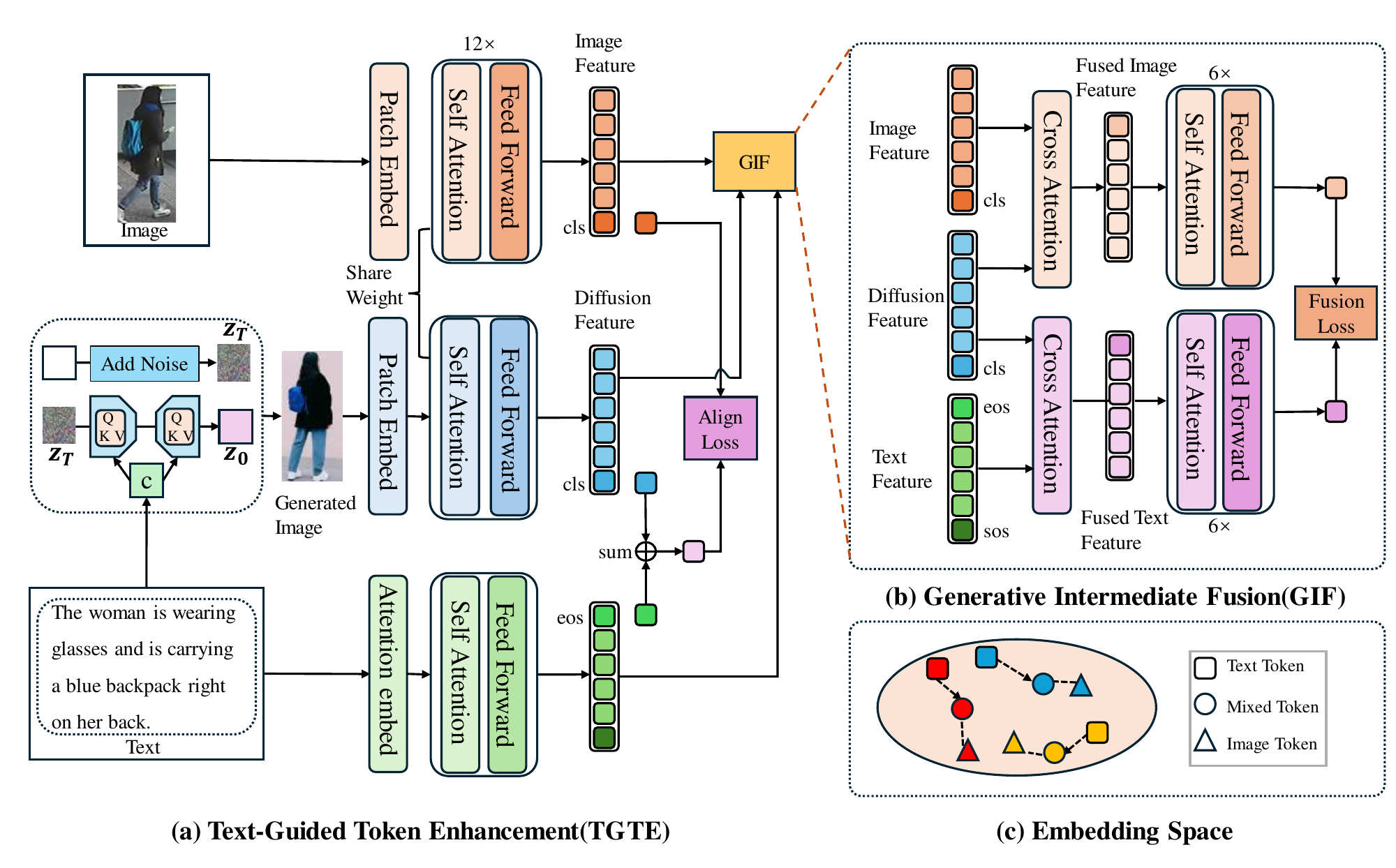}
\vspace{0.5em} 
\captionsetup{justification=justified, singlelinecheck=false}
\caption{
Framework of the GEA method. TGTE is responsible for generating and encoding intermediate images and extracts features from the input text, original image, and a diffusion-generated intermediate image, while GIF facilitates cross-modal interaction and implicit relational inference across the three representations via cross fusion. (a) is the illustration of the TGTE, which shows three parallel branches that extract features from the original image, the input text, and a diffusion-generated image conditioned on the text prompt. The generated image acts as an intermediate modality to enrich textual semantics. (b) illustrates the Generative Intermediate Fusion (GIF) module, where cross-attention mechanisms align the generated features with the original image and text features, respectively. (c) displays the core idea of our approach with three identities, marked in red, blue, and yellow. Our method pulls matched image-text features closer and separates non-matching ones. 
}

\label{fig:framework}
\vspace{0.8em} 
\end{figure*}

\subsection{Text Guided Diffusion Model}
Due to the problems of pattern collapse and unstable training in generative adversarial network (GANS)~\cite{goodfellow2014generative}.
The diffusion model~\cite{ho2020denoising} introduces noise and the denoising process to obtain more stable training and high-fidelity output. The denoising diffusion probability model (DDPM)~\cite{ho2020ddpm} significantly improves the quality and efficiency of generation, especially in large-scale and multimodal environments.
Recent advances in text-guided image generation have been largely driven by diffusion-based models, among which Stable Diffusion~\cite{rombach2022high} has achieved remarkable success. By leveraging cross-attention mechanisms to integrate textual embeddings during the denoising process, Stable Diffusion enables precise and semantically aligned image synthesis from natural language prompts. Its outstanding ability to produce high-resolution, visually realistic, and semantically consistent images has established it as one of the most effective solutions for text-to-image generation. The latest iteration, the Stable Diffusion 3 (SD-3)~\cite{esser2024scaling}, further improves semantic understanding and generative quality, making it a powerful and ready-to-use tool for tasks that require visually grounded representations guided by textual input.

Beyond image generation, diffusion-based frameworks have demonstrated remarkable versatility in a variety of complex multimodal tasks~\cite{bao2023one,couairon2022diffedit}.
Recent works have successfully explored diffusion models for retrieval tasks, demonstrating their powerful ability to align semantic features across modalities~\cite{jin2023diffusionret,long2025zero, tang2023emergent}.
However, in Text-to-Image Person Retrieval (TIPR) tasks, a major challenge lies in the inherent incompleteness and abstraction of textual descriptions. 
Fortunately, Stable Diffusion~\cite{rombach2022high} have shown strong capabilities in generating images that faithfully reflect and even amplify the semantic information present in the input text.
By conditioning on sparse textual prompts, diffusion models can synthesize detailed visual content that enhances the semantic expressiveness of the original descriptions.
In this work, we adopt a new perspective that utilizes the rich semantic priors encoded in diffusion-generated images as intermediate representations for cross-modal alignment. 


\section{Method}
In this section, we will elaborate on the GEA method. The overall architecture of our proposed framework is illustrated in Figure~2. It consists of three core components:  
(1) a backbone network for feature extraction from both image and text,  
(2) a Text-Guided Token Enhancement (TGTE) module that produces semantically enriched images from textual descriptions,  
(3) a Generative Intermediate Fusion (GIF) that refines local correspondences across three modalities and enhance fine-grained cross-modal semantic consistency.
In Section 3.1, we will introduce the backbone structure. In Section 3.2, we will introduce the TGTE module. Finally, in Section 3.3, we will introduce the GIF module, which facilitates effective cross-modal alignment. The algorithm process is in supplementary material.
\subsection{Backbone Network}
We adopt CLIP~\cite{radford2021learning} as the backbone for both image and text encoders in our GEA framework. CLIP is a powerful vision-language pretraining model trained on a large-scale dataset of image-text pairs using a contrastive learning objective.  To obtain token representations and implement cross-modal interactions, it learns to project images and texts into a shared embedding space through two embedding modules. In the embedding space, semantically aligned image-text pairs have high similarity. Due to its strong cross-modal alignment capability and generalization performance, CLIP serves as an effective foundation for TIPR tasks.

\textbf{Image Encoder.}
For an input image $ I_i \in R^{H \times W \times C} $. The ViT encoder divides the input image into $ M = \frac{H \times W}{P^2} $ non-overlapping patches with positional information, where P is the patch size. These patches will be input into the Transformer network and processed through multiple layers of the Transformer. After multiple self-attention layers and linear projection, the global contextual information of the image is extracted. At a result, the image encoder outputs a token vector $\{\textbf{v}_{cls}, \textbf{v}_1, \dots, \textbf{v}_M\}$ , in which the $ \textbf{v}_{\text{cls}} $ token serves as the global image representation.

\textbf{Text Encoder.}
The input text $T_j$ is first tokenized through a lower-cased Byte Pair Encoding (BPE)~\cite{sennrich2016bpe} with a vocabulary size of 49,152. After processing by the CLIP text encoder, which consists of a stack of transformer~\cite{vaswani2017attention} layers with multi-head self-attention. We get the output token list $ \{\textbf{t}_{sos}, \textbf{t}_1, \dots, \textbf{t}_N,\textbf{t}_{eos}\} \ $. In the final transformer layer, the embedding corresponding to the $\textbf{t}_{\text{eos}}$ token is selected as the global representation of the entire textual input.

\subsection{Text-Guided Token Enhancement}

As mentioned before, textual descriptions are sometimes incomplete and may omit critical visual details. To enhance the expressive capacity of sparse textual descriptions, we propose the Text-Guided Token Enhancement (TGTE) module.

TGTE leverages diffusion-generated intermediate images to enrich and enhance the semantic expression of sparse text inputs. 
The generated image is encoded into feature embeddings and fused with the original text features, resulting in a more semantically enriched representation. 
Importantly, this process amplifies the informative aspects already present in the text without introducing unverifiable or hallucinated content, thereby enhancing the reliability of the cross-modal alignment.
By strengthening the existing textual semantics, TGTE facilitates more accurate and robust alignment between text and visual modalities, thus narrowing the modality gap in TIPR tasks.

Given a text query, we employ a pretrained Stable Diffusion 3-medium model~\cite{esser2024scaling} to generate an intermediate image conditioned on the text prompt.
Stable Diffusion~3 is a latent rectified-flow model with a Multimodal Diffusion Transformer (MMDiT) backbone that performs denoising in a compressed latent space, enabling efficient high-resolution generation. 
During inference, text embeddings (from pretrained text encoders) modulate transformer blocks and are jointly processed with latent image tokens via the MMDiT's bidirectional attention, guiding the generation towards a semantically consistent image. 
The architecture comprises three core components:

\begin{itemize}
    \item \textbf{MMDiT Denoiser:} A multimodal diffusion transformer operates in latent space and iteratively refines a noisy latent $z_t$ conditioned on the text tokens.
    \item \textbf{Text Encoders:} Pretrained text encoders convert the text description into conditioning embeddings that modulate the transformer via attention and normalization layers.
    \item \textbf{VAE Decoder:} A variational autoencoder (VAE) $D$ decodes the final latent back into pixel space to obtain the image.
\end{itemize}

We start from a standard Gaussian latent $z_1 \sim \mathcal{N}(0, I)$ in the VAE latent space and solve a rectified-flow ODE from $t{=}1$ to $t{=}0$. 
At each (discretized) step, the model predicts a velocity field $v_\theta$ conditioned on the current latent and text embedding $c$, and we update
\begin{equation}
\frac{dz_t}{dt} \;=\; -\,v_\theta(z_t, t, c), \qquad
z_{t-\Delta t} \;=\; z_t \;-\; \Delta t \, v_\theta(z_t, t, c).
\end{equation}
After integrating to $t{=}0$, the final latent $z_0$ is decoded by the VAE to produce the synthetic image:
\begin{equation}
G \;=\; D(z_0).
\end{equation}

After obtaining the generated image $G$, we pass the text $T$, image $I$, and generated image $G$ through CLIP text/image encoders to get the global representation $\textbf{t}_{eos}$, $\textbf{v}_{cls}$ and $\textbf{g}_{cls}$ in a shared feature space. Notably, the generated image is not treated as a retrieval candidate but as an intermediate representation that provides semantically rich, text-aligned visual information. This intermediate representation enhances the expressive power of the textual modality and facilitates more robust cross-modal fusion. We reformulate the text feature as the mix of $\textbf{t}_{eos}$ and $\textbf{g}_{cls}$. We name the new token as $\textbf{t}_{cls}$ and balance the contributions of visual and textual embeddings via a weight parameter $\omega$. The token $\textbf{t}_{cls}$ can be expressed using the following formula:
\begin{equation}
\textbf{t}_{cls} = (1-\omega) \textbf{t}_{eos} +\omega \textbf{g}_{cls}.
\end{equation}

When obtaining the global tokens for the image ($\textbf{v}_{cls}$) and the mix of generated image and text ($\textbf{t}_{cls}$) respectively, we compute their similarity to estimate the matching probability.
For each candidate image $I_i \in I$ and text query $T_j \in T$, the cosine similarity score between $(I_i, T_j)$ is calculated as:
\begin{equation}
 S(I_i, T_j) = \frac{\textbf{v}^i_{cls}  \cdot  \textbf{t}^j_{cls}}{\left\| \textbf{v}^i_{cls} \right\| \cdot \left\|\textbf{t}^j_{cls} \right\|}.
\end{equation}

To train the model for better cross-modal alignment, we optimize the feature representations such that the matched image-text pairs have higher similarity scores than mismatched ones. 
In this work, we adopt the triplet alignment loss (TAL)~\cite{Qin_2024_CVPR} to supervise the learning process. For a pair of image-text sample ($I_i$, $T_i$) in a batch of $K$ samples, the align loss can be formulated as:
\begin{align}
  \mathcal{L}_{Align}(I_i,T_i)=[m-S^+_{i2t}(I_i)+\tau \text{log}\sum\limits_{j=1}^Kexp(S(I_i,T_j)/\tau]_+\notag \\+[m-S^+_{t2i}(T_i)+\tau \text{log}\sum\limits_{j=1}^Kexp(S(I_j,T_i)/\tau]_+,
\end{align}
where \emph{m} is a positive margin coefficient,$\tau$ is a temperature coefficient to control hardness. $[x]_+ = max(x,0),exp(x) = e^x$, \emph{K} is the size of the input data in a batch. $S^+_{i2t}(I_i)=\sum\limits_{j=1}^K\alpha_{ij}S(I_i,T_j)$, where $\alpha_{ij}$ is defined as:
\begin{equation}
\alpha_{ij}=\frac{exp(S(I_i,T_j)/\tau)}{\sum\limits_{n=1}^Kexp((S(I_i,T_n)/\tau)},
\end{equation}
where $S^+_{t2i}(T_i)$
is the same weighted average similarity of positive pairs for text $T_i$.


\subsection{Generative Intermediate Fusion}
To better align multimodal features, we introduce the Generative Intermediate Fusion (GIF) module, as shown in Figure~\ref{fig:framework}(b). 
GIF integrates information from three modalities: the original image, the input text, and the diffusion-generated image. 
GIF leverages the generated image as a semantic enhancer to facilitate fine-grained alignment between text and visual representations. Even if the input description is incomplete, this fusion strategy is still effective because the generated image always corresponds to the original text description, serving as an intermediary to help with better fine-grained alignment.

GIF employs two independent cross-attention modules: one aligns the generated image with the original image, and the other aligns it with the text. 
The output features are then fused through a transformer encoder, producing the fused representations $\textbf{v}^f$ and $\textbf{t}^f$, which capture subtle semantic correspondences without introducing unverifiable information. 
In practice, each branch of GIF consists of a Cross-Attention (CA) layer followed by a 6-layer transformer block, ensuring sufficient capacity for fine-grained semantic reasoning. The feature processing between images and text can be expressed in the following ways:

\begin{equation}
 \ \textbf{v}^f=Transformer(CA(LN(Q,K,V))),\
\end{equation}
\begin{equation}
 \ \textbf{t}^f=Transformer(CA(LN(Q,K,V))),\
\end{equation}
where LN(·) represent Layer Normalization, CA(·) is defined as follows:
\begin{equation}
 CA(Q,K,V)=softmax(\frac{QK^T}{\sqrt{d}})V,
\end{equation}
For Cross-Attention layers, we set the generated image feature as K and V and set text or image feature as Q, where d is the embedding dimension of the tokens before fusion.
Finally, a Triplet Alignment Loss (TAL) is applied to calculate the fusion loss$\mathcal{L}_{Fusion}$. The calculation of TAL is already explained in Section 3.2.
The final training objective is the sum of the global alignment loss and the fusion alignment loss:
\begin{equation}
 \mathcal{L} = \mathcal{L}_{Align} + \mathcal{L}_{Fusion}
\end{equation}

\begin{table*}[h]
    \centering
    \caption{Comparison with state-of-the-art methods on CUHK-PEDES, RSTPReid, and ICFG-PEDES datasets. 
    R-1,5,10 is an abbreviation for Rank-1,5,10\% accuracy. 
    The best and second-best results are in \textbf{bold} and \underline{underline}, respectively.}
    \label{tab:sota_comparison}
    \resizebox{\textwidth}{!}{
    \begin{tabular}{l|l|cccc|cccc|cccc}
        \toprule
        \multirow{2}{*}{\textbf{Method}} & \multirow{2}{*}{\textbf{Reference}} 
        & \multicolumn{4}{c|}{\textbf{CUHK-PEDES}} 
        & \multicolumn{4}{c|}{\textbf{RSTPReid}} 
        & \multicolumn{4}{c}{\textbf{ICFG-PEDES}} \\
        & & R-1 & R-5 & R-10 & mAP & R-1 & R-5 & R-10 & mAP & R-1 & R-5 & R-10 & mAP \\
        \midrule
        SSAN~\cite{ding2021semantically} & arXiv'21 & 61.37 & 80.15 & 86.73 &-  & 43.50 & 67.80 & 77.15 & - & 54.23 & 72.63 & 79.53 &-  \\
        IVT~\cite{shu2022see} & ECCVW'22 & 65.59 & 83.11 & 89.21 &-  & 46.70 & 70.00 & 78.80 & - & 56.04 & 73.60 & 80.22 &-  \\
        CFine~\cite{yan2023clip} & TIP'23 & 69.57 & 85.93 & 91.15 &-  & 50.55 & 72.50 & 81.60 & - & 60.83 & 76.55 & 82.42 &-  \\
        IRRA~\cite{Jiang_2023_CVPR} & CVPR'23 & 73.38 & 89.93 & 93.71 & 66.13 & 60.20 & 81.30 & 88.20 & 47.17 & 63.46 & 80.25 & 85.82 & 38.06 \\
        RaSa~\cite{bai2023rasa} & IJCAI'23 & \underline{76.51} & {90.29} & {94.25} & \underline{69.38} & {66.90} & \underline{86.50} & {91.35} & {52.31} & 65.28 & 80.40 & 85.12 & \underline{41.29} \\
        CSKT~\cite{liu2024clip} & ICASSP'24 & 69.70 & 86.92 & 91.80 & 62.74 & 57.75 & 81.30 & 88.35 & 46.43 & 58.90 & 77.31 & 83.56 & 33.87 \\
        FLAN~\cite{xie2024full} & ESWA'24 & 71.89 & 88.54 & 93.28 & - & 57.79 & 80.82 & 87.45 & - & 61.39 & 78.65 & 84.58 & - \\
        EAIBC~\cite{zhu2024improving} & TNNLS'24 & 64.96 & 83.36 & 88.42 & - & 49.85 & 70.15 & 79.85 &-  & 58.95 & 75.95 & 81.72 &-  \\
        RDE~\cite{Qin_2024_CVPR} & CVPR'24 & 75.94 & 90.14 & 94.12 & 67.56 & 65.35 & 83.95 & 89.90 & 50.88 & \underline{67.68} & {82.47} & {87.36} & 40.06 \\
        AMNS~\cite{zhang2025amns} & ICASSP'25 & 74.75 & 89.54 & 93.83 & 67.67 & 60.50 & 81.30 & 87.65 & 47.20 & 64.05 & 79.90 & 85.90 & 41.27 \\
        MMRef~\cite{ma2025multi} & TMM'25 & 72.25 & 88.24 & 92.61 & - & 56.20 & 77.10 & 85.80 &-  & 63.50 & 78.19 & 83.73 & - \\
        ICL~\cite{Qin_2025_CVPR} & CVPR'25 & {76.41} & \underline{90.48} & \underline{94.33} & {68.04} & \textbf{67.70} & {86.05} & \underline{91.75} & \underline{52.62}  & \textbf{68.11} & \underline{82.59} & \underline{87.52} & {40.81} \\
        \textbf{Ours (GEA)} & - & \textbf{80.56} & \textbf{93.37} & \textbf{96.44} & \textbf{72.73} & \underline{67.60} & \textbf{87.50} & \textbf{93.35} & \textbf{54.03} & {65.56} & \textbf{82.70} & \textbf{87.88} & \textbf{43.08} \\

        \bottomrule
    \end{tabular}}
\end{table*}


\section{Experiments}
We conduct extensive experiments on three challenging Text-to-Image Person Retrieval datasets: CUHK-PEDES~\cite{li2017cuhkpedes}, RSTPReid~\cite{zhu2021dssl}, and ICFG-PEDES~\cite{ding2021semantically}. These datasets are widely used in TIPR research and serve as standard benchmarks for evaluating text-to-image retrieval performance in diverse scenarios. 
\subsection{Datasets and Settings}
In this section, we introduce the datasets used for evaluation, the adopted evaluation metrics, and the implementation details of our experimental setup. 
Our evaluation protocol follows the common practice in prior TIPR studies, ensuring a fair and consistent comparison with existing state-of-the-art methods. 


\subsubsection{Datasets}
CUHK-PEDES, RSTPReid, and ICFG-PEDES are widely used TIPR datasets that contain numerous person images and associated natural language descriptions.
Following IRRA~\cite{Jiang_2023_CVPR}, we use the standard data splits for training, validation, and testing. 
For ICFG-PEDES, since no official test set is provided, we use the validation set for evaluation. 
\subsubsection{Evaluation Metrics}
We adopt the popular Rank-k metrics (k = 1, 5, 10) as standard evaluation measures for retrieval performance. It represents the proportion of queries for which the correct image appears within the top-k retrieved results.

In addition to Rank-k, we take the mean Average Precision (mAP) into consideration, which evaluates both the correctness and ranking quality of the retrieved results. mAP is computed by taking the average of precision values across all correct matches for each query, then averaging across all queries.

\subsubsection{Implementation Details}
As mentioned earlier, we utilize a pre-trained CLIP~\cite{radford2021learning} model as our encoder, along with two randomly initialized multimodal interactive encoders. In fairness, we use the same version of CLIP-ViTB/16 as IRRA~\cite{Jiang_2023_CVPR}
to conduct experiments. Each hidden layer has a dimension of 512 and employs 8 attention heads.  
We apply random horizontal flipping, random cropping, and random erasing for data augmentation. All images in the dataset are resized to \(384 \times 128\) $(H \times W)$, with a batch size of 64. Each image is divided into 192 $(M)$ patches with a shape of \(16 \times 16\) $(P \times P)$. The token length for text inputs is fixed at 77 $(N)$.  
The model is trained for 60 epochs using the Adam optimizer with a pre-learning rate decay strategy. The initial learning rate for CLIP is set to \(1 \times 10^{-5}\), while the learning rate for the fusion module is \(1 \times 10^{-4}\). We incorporate a 5-epoch warm-up phase, where the learning rate is linearly increased from \(1 \times 10^{-6}\) to \(1 \times 10^{-5}\).  
Following RDE~\cite{Qin_2024_CVPR}, for the hyperparameters of TAL, we set the margin value \(m\) to 0.1 and the temperature parameter \(\tau\) to 0.015.
We use the pretrained Stable Diffusion 3‑medium model as our generator with 28 inference steps. For each text description, we use the original text as the positive prompt and append “pedestrian” to reinforce the person-centric semantics, while using “cartoon” as a negative prompt to suppress cartoon-style artifacts. Generated images are produced at a resolution of 1024 $\times$ 336 with the guidance scale set to 7, and only one image is generated per text. The mix weight $w$ gradually increases from 0.3 to 0.6.
We conduct relevant experiments on a single RTX4090 24GB GPU. 

\subsection{Comparison with State-of-the-Art Methods}
In this section, we compare our proposed method with several state-of-the-art approaches to evaluate its effectiveness on cross-modal person retrieval tasks. We select a series of representative baselines, including traditional methods based on global feature matching, recent approaches employing fine-grained alignment, and advanced models leveraging pre-trained backbones such as CLIP.

As shown in Table 1, our method achieves competitive or superior performance across most evaluation metrics and datasets. Compared with the most recent state-of-the-art methods, our model achieves strong performance across metrics in R-1, R-5, R-10, and mAP on both CUHK-PEDES and RSTPReid datasets. 

On the CUHK-PEDES dataset, our model achieves an R-1 score of 80.56\% and mAP of 72.73\%, outperforming the previous best method Rasa (76.51\% / 69.38\%) by +4.05\% in R-1 and +3.35\% in mAP.
On the RSTPReid dataset, our method obtains 67.60\% on R-1 and 54.03\% on mAP, slightly behind ICL in R-1 (67.70\%) but surpassing it in mAP by +1.41\%. 
On the ICFG-PEDES dataset, our method performs competitively. It achieves the highest mAP score of 43.08\% and ranks among the top in R-5 and R-10. The R-1 score (65.56\%) is lower than ICL (68.11\%). This is primarily due to the high level of textual noise and ambiguity in the ICFG annotations, which limits the accuracy of our generated images. Since the intermediate images are generated based on the textual prompts, noisy or ambiguous descriptions may lead to less precise bridging representations and affect top-1 matching. Nevertheless, the significant improvements in R-5, R-10, and mAP indicate that our method remains effective in improving the overall retrieval ranking, particularly by enhancing semantic alignment on a broader candidate scale.

In summary, the strong performance of our proposed GEA framework across multiple datasets and evaluation metrics demonstrates the effectiveness of our overall design. The consistent improvements in R-1, R-5, R-10, and mAP validate the contributions of each key component, particularly the diffusion-enhanced semantic generation and cross-modal fusion module.

\textbf{Discussion: Why does GEA outperform?}
The consistent performance gains of GEA can be attributed to two key factors. 
First, the \textbf{Text-Guided Token Enhancement (TGTE)} enriches the semantic representation of sparse textual descriptions by generating intermediate images conditioned on text prompts. 
Rather than augmenting the dataset, these generated images enhance the expression of limited semantic cues—such as clothing texture, color, or accessories—providing more informative visual anchors for alignment without altering the retrieval targets.
Second, the \textbf{Generative Intermediate Fusion (GIF)} module effectively integrates features from the original image, the generated image, and the text using dual cross-attention and a fusion transformer. 
GIF focuses the model’s attention on fine-grained, text-relevant visual details.
TGTE and GIF form a generative-enhanced retrieval framework that narrows the modality gap and strengthens fine-grained alignment, leading to superior retrieval performance across diverse datasets.

\subsection{Ablation Study}

In this section, we conduct our ablation experiments on the CUHK-PEDES dataset to explore the roles of different methods and modules in GEA. As shown in Table 2, we conduct an ablation study to investigate the individual contributions of the TGTE and the GIF.

By comparing Row \#1 and Row \#3, we observe that incorporating TGTE leads to a substantial performance improvement, with R-1 increasing from 70.84\% to 78.57\%, and mAP rising from 62.54\% to 70.89\%. It is important to note that these generated images are not used as additional training data or external augmentation; instead, they serve as intermediate semantic representations during inference, derived from the input text via a pretrained diffusion model. This highlights the unique role of diffusion in transforming abstract textual descriptions into visually grounded features, thereby enriching the text representation and narrowing the modality gap between text and image.

Further comparing Row \#3 and Row \#4, we see that the introduction of our GIF module yields further gains (R-1 from 78.57\% to 80.41\%, mAP from 70.89\% to 72.86\%). This confirms the effectiveness of our cross-modal fusion strategy, which explicitly integrates original visual features, textual features, and diffusion-generated features in a unified representation space. Unlike traditional alignment methods that rely solely on patch-word matching or global feature similarity, our approach introduces a semantically aligned intermediate modality that bridges the inherent gap between text and image modality.

The GIF module operates as a semantic aggregator, enabling bidirectional information flow between the modalities through cross-attention. This design allows the model to selectively focus on complementary cues, such as details missed in the text but captured in the generated image, thereby achieving more precise cross-modal alignment. The consistent improvements in both ranking accuracy and mAP across settings validate the effectiveness of our generative-enhanced alignment mechanism and demonstrate the practical value of using synthetic semantics to support real-world retrieval tasks.

On the other hand, comparing Row \#4 and Row \#3 shows that the GIF module alone brings limited improvement when not supported by generative features, highlighting that the TGTE module are crucial for bridging the modality gap.

These results validate the complementary effects of our two core components and confirm the overall effectiveness of our proposed GEA framework.

\newcolumntype{Y}[1]{>{\centering\arraybackslash}m{#1}}  

\begin{table}[h]
    \centering
    \renewcommand{\arraystretch}{1.4}
    \vspace{0.2cm}
    \caption{Ablation study on TGTE and GIF}
    \label{tab:ablation_study}
    \begin{tabularx}{\linewidth}{
        Y{0.8cm}  
        Y{0.8cm}  
        Y{0.8cm}  
        Y{0.8cm}  
        Y{0.8cm}  
        Y{0.8cm}  
        Y{0.8cm}  
    }
        \toprule
        \textbf{Model} & \textbf{TGTE} & \textbf{GIF} & \textbf{R-1} & \textbf{R-5} & \textbf{R-10} & \textbf{mAP} \\
        \midrule
        \#1 & --         & --          & 70.84         & 87.44         & 92.12         & 62.54 \\
        \#2 & --         & \checkmark & 72.82         & 89.75         & 93.78         & 64.69 \\
        \#3 & \checkmark & --          & 78.57         & 92.82         & 96.18         & 70.89 \\
        \#4 & \checkmark & \checkmark & \textbf{80.56} & \textbf{93.37} & \textbf{96.44} & \textbf{72.73} \\
        \bottomrule
    \end{tabularx}
\end{table}

\subsection{Visualization Analysis}
To further validate the effectiveness of our proposed method from a qualitative perspective, we conduct visual analyses to illustrate how our GEA framework enhances cross-modal alignment and fine-grained semantic understanding.

\subsubsection{Qualitative Analysis via t-SNE}
In the ablation experiments, we observed that TGTE played a crucial role in improving model performance by enriching the semantic expression of text features.

\begin{figure*}[t]
\centering
\includegraphics[width=16cm]{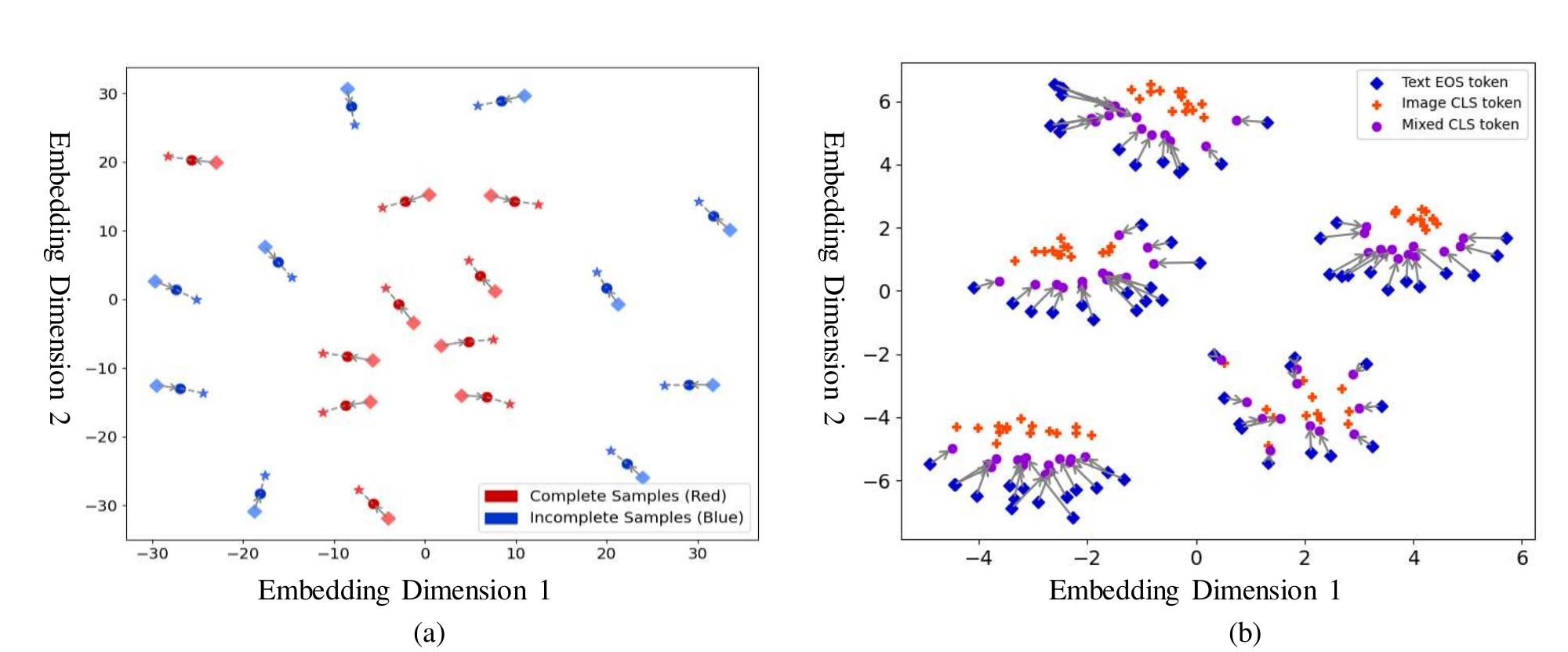}
\caption{t-SNE visualization of feature embeddings. (a) Visualization of 10 samples with complete textual descriptions and 10 with incomplete descriptions. The fused tokens (circle) consistently lie between their corresponding text (square) and image tokens (star), indicating our method’s ability to bridge the modality gap regardless of text completeness.  
(b) Visualization of all 67 samples from five randomly selected identities. Our method effectively clusters corresponding image-text pairs while separating unrelated samples. }
\label{fig:tsne}
\vspace{0.5cm} 
\end{figure*}

To further verify this effect, we visualize the feature distributions using t-SNE. As shown in Figure~\ref{fig:tsne} (a), We selected 10 samples with incomplete text descriptions and 10 samples with complete descriptions. 
We project the [CLS] tokens from the image encoder, the [EOS] tokens from the text encoder, and the fused [CLS] features into a 2D space.
As illustrated, regardless of whether the text is complete or incomplete, the fused features consistently lie between their corresponding text and image features.
This spatial relationship shows that our fusion effectively serves as a semantic bridge, pulling heterogeneous features closer together in the embedding space.
Such tighter and more coherent distributions validate our core motivation: using generative-enhanced representations to mitigate modality discrepancies and strengthen cross-modal alignment.
We also visualize all samples from five randomly selected identities using t-SNE. As shown in Figure~\ref{fig:tsne} (b), our method successfully clusters matching image-text pairs more tightly while pulling apart unrelated ones, confirming its effectiveness in improving cross-modal discriminability.


\subsubsection{Attention Heatmap Analysis}
\begin{figure}[t]
\centering
\includegraphics[width=8cm]{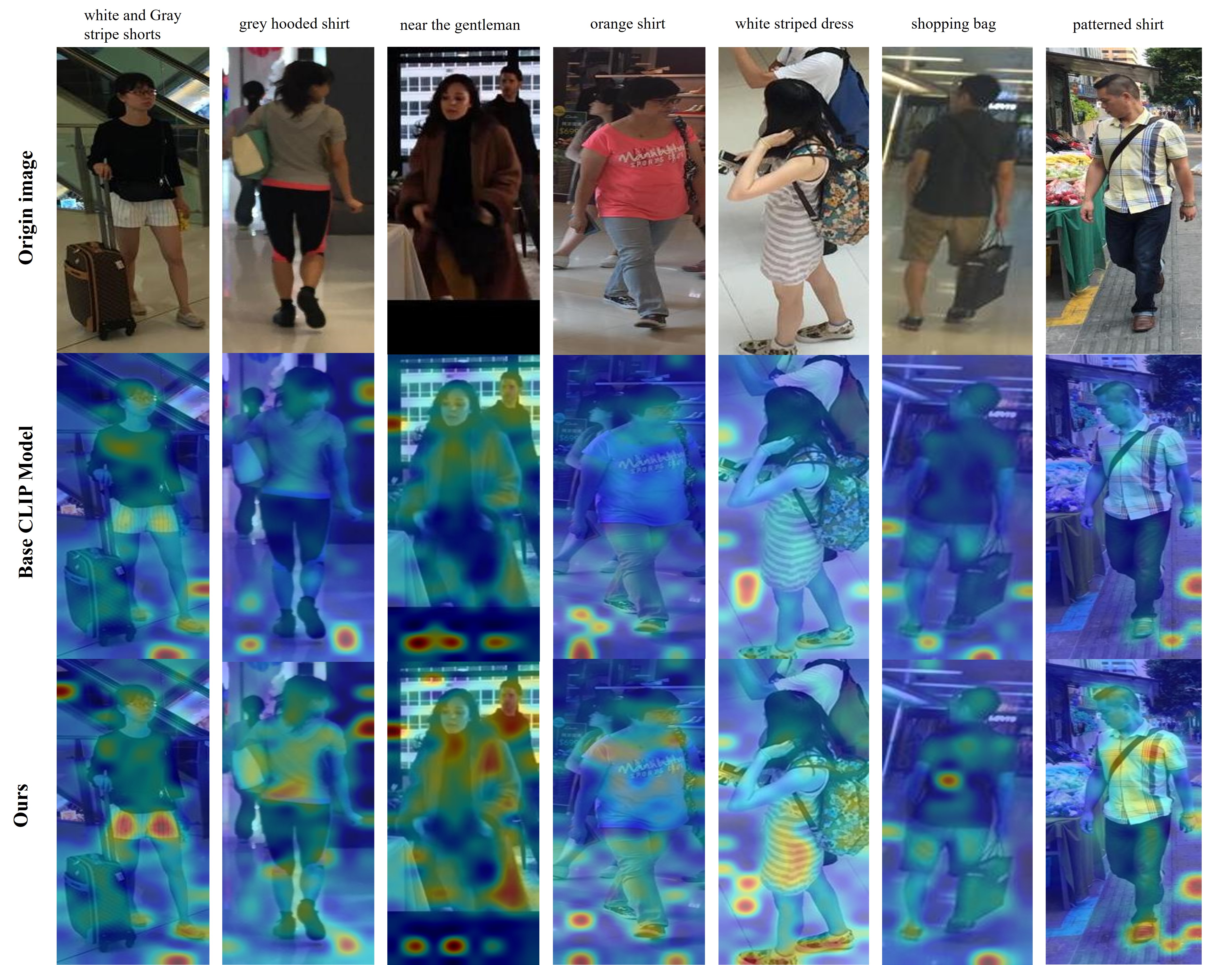}
\caption{Attention heatmap visualization comparing the baseline CLIP model and our proposed method. For each input pair, the top row shows the original image and its corresponding text description. The middle row displays the attention map from the baseline model, while the bottom row shows the attention map produced by our method.}
\label{fig:heatmap}
\vspace{0.5cm} 
\end{figure}

In cross-modal tasks, there are often modal differences between text and image features, and direct alignment may result in information loss or matching bias. GIF introduces a fusion mechanism to better integrate them in the semantic space.
To verify our GIF module effectively focuses on fine-grained semantic features, we employ attention response heatmaps to provide visual evidence. We visualize the attention maps over the image regions in response to corresponding textual descriptions. Figure~\ref{fig:heatmap} presents attention heatmaps that compare the Base CLIP model and our proposed method across several text-image pairs. 

We observe that the Base CLIP model often fails to localize key visual cues described in the text. The contrast between Line 2 and Line 3 clearly demonstrates that our model accurately directs attention to the corresponding semantic regions described in the sparse text like "near the gentleman", "patterned shirt", and "white striped dress". This indicates that our method can effectively catch the fine-grained semantic details faithfully aligned even with the sparse textual input.
This improvement stems from our fusion mechanism, which leverages the diffusion-generated image as an intermediate modality, facilitates mutual perception between the original image and the text. This enables the model to better capture fine-grained correspondences between modalities.


\subsection{Parametric Analysis}
\begin{figure}[t]
\centering
\includegraphics[width=8cm]{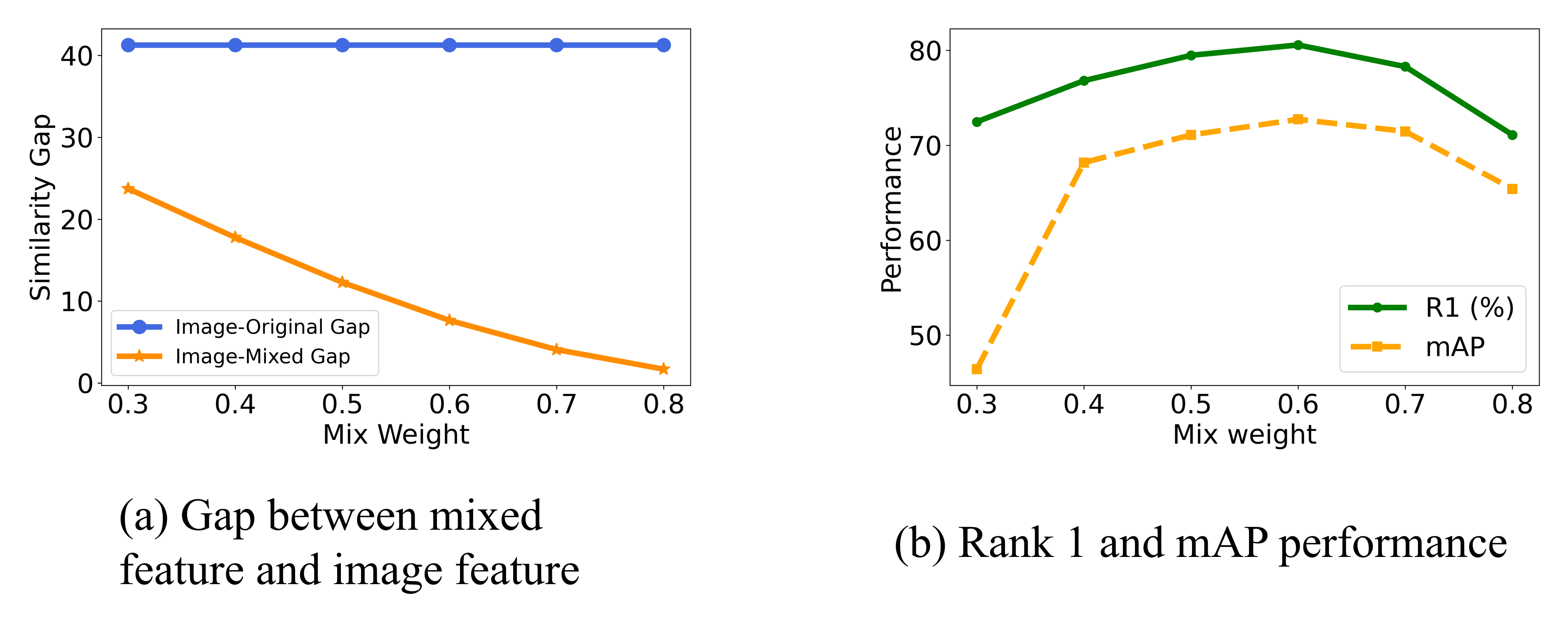}
\caption{Variation of performance with different fusion weight.}
\label{fig:para}
\vspace{0.5cm}
\end{figure}

Although incorporating generated images as a supplement to text features can enhance semantic representation, the impact of the mix weight on overall performance remains unclear. We conduct a sensitivity analysis of the mix weight parameter $w$ on the CUHK-PEDES dataset. As shown in the figure~\ref{fig:para}, we observe two main trends: (1) the mix weight increases, the similarity differences between positive pairs tend to converge; (2) when the mix weight becomes too large, it adversely affects the model’s performance, leading to a noticeable drop in retrieval accuracy.

\section{Limitations and Future Work}
Although our proposed GEA framework demonstrates strong performance across multiple TIPR benchmarks, it still faces some limitations. Specifically, when the textual descriptions contain noise, the generated images may inadvertently amplify these deficiencies. In addition, if the color of a specific body region is not specified in the input text, the generated result is unconstrained and may differ from the original corresponding image. As a result, the modality gap may persist or even increase, affecting retrieval accuracy in such scenarios.
In future work, we plan to explore strategies for detecting and filtering noisy or ambiguous text queries before generation. 
This could involve developing confidence-aware or uncertainty-aware generation mechanisms that adaptively assess the reliability of the input text and adjust the generation process accordingly. 

\section{Conclusion}

In this paper, we reveal two key challenges in Text-to-Image Person Retrieval (TIPR): the incompleteness of textual descriptions and the limited size and diversity of existing datasets.  
To this end, we propose Generation-Enhanced Alignment (GEA), a generative-based framework that handles the problems mentioned before.  
Extensive experiments on three benchmark datasets demonstrate the effectiveness of our method.

\begin{ack}
This work was supported by the Natural Science Foundation of China (No.62372082), Shenzhen Natural Science Foundation  (No.JCYJ20240813114206010) and the Fundamental Research Funds for the Central Universities (No. ZYGX2024Z017).
\end{ack}

\bibliography{mybibfile}   


\end{document}